\documentclass[letterpaper, 10 pt, journal, twoside]{IEEEtran}

\IEEEoverridecommandlockouts 

\usepackage{graphics} 
\usepackage{epsfig} 
\usepackage{mathptmx} 
\usepackage{times} 
\usepackage{amsmath} 
\usepackage[T1]{fontenc}
\usepackage{amssymb} 

\usepackage{placeins}

\newcommand{\norm}[1]{\left\lVert#1\right\rVert}

\title{Geometrically Mappable Image Features} 

\author{Janine Thoma, Danda Pani Paudel, Ajad Chhatkuli, and Luc Van Gool%
\thanks{Manuscript received September 10, 2019; Revised December 8, 2019;
Accepted January 8, 2020.} 
\thanks{This paper was recommended for publication by Editor Sven Behnke upon
evaluation of the Associate Editor and Reviewers' comments. This research has received funding from Huawei Gift for research and the EU Horizon 2020 research and innovation program under grant agreement No.\ 820434.} 
\thanks{J.~Thoma, D.~Paudel, and A.~Chhatkuli are with the Computer Vision Lab, ETH, Zurich 8092, Switzerland (e-mail: \{jthoma, paudel, ajad.chhatkuli\}@vision.ee.ethz.ch).}%
\thanks{L.~Van Gool is with the Computer Vision Lab, ETH, Zurich 8092, Switzerland, and also with the Katholieke Universiteit Leuven, Leuven 3000, Belgium
(e-mail: vangool@vision.ee.ethz.ch).}%
\thanks{Digital Object Identifier (DOI): see top of this page.}
}

\begin{document}

\maketitle

\begin{abstract}
Vision-based localization of an agent in a map is an important problem in robotics and computer vision. In that context, localization by learning matchable image features is gaining popularity due to recent advances in machine learning. Features that uniquely describe the visual contents of images have a wide range of applications, including image retrieval and understanding. In this work, we propose a method that learns image features targeted for image-retrieval-based localization. Retrieval-based localization has several benefits, such as easy maintenance and quick computation. However, the state-of-the-art features only provide visual similarity scores which do not explicitly reveal the geometric distance between query and retrieved images. Knowing this distance is highly desirable for accurate localization, especially when the reference images are sparsely distributed in the scene. Therefore, we propose a novel loss function for learning image features which are both visually representative and geometrically relatable. This is achieved by guiding the learning process such that the feature and geometric distances between images are directly proportional. In our experiments we show that our features not only offer significantly better localization accuracy, but also allow to estimate the trajectory of a query sequence in absence of the reference images.  
\end{abstract}


\section{Introduction}
Navigation is a fundamental task in robotics which needs to be solved by autonomous agents.
Vision-based navigation has the potential to offer accurate and robust localization where GPS information and GPS-tagged maps are not reliable or entirely unavailable. Applications range from space exploring robots and self-driving cars to the more mundane task of aiding people with navigation using mobile devices.
A key problem of vision-based navigation is to localize the robot or agent in a map given any image(s) of the physical environment. This is a challenging task, as the scene may contain dynamic elements and may undergo severe seasonal and day-night changes. 

We are concerned with the latter. The problem of image-retrieval-based localization involves learning a feature embedding from a sparse set of geo-tagged images, also called the reference or landmark images. These features---along with their geo-location---form the map. We then compute the visual feature of a query image taken by the agent and compare it with the map feature embedding, finally achieving localization by nearest neighbor search in the $\ell_2$-norm sense. Such simplicity of image-retrieval-based localization results in speed and robustness which is difficult to achieve with SLAM or SfM based methods. This is because the latter rely entirely on sparse, noisy and outlier-prone local feature descriptors. These local feature descriptors may undergo large changes with dynamic, seasonal and day-night changes of the environment. The way to reliably address such issues is to incorporate the changes in the map itself which may not always be feasible due to computational and memory restrictions.
An alternative approach to SLAM or SfM for 6-DoF pose estimation is regression through CNNs~\cite{Walch_2017_ICCV, Kendall2015, Kendall2017, Taira2018}. These methods learn to encode object poses from images using the ground truth 6-DoF camera poses. Although interesting and powerful in theory, such approaches require ground truth 6-DoF pose, making the training data scarce and difficult to acquire. The data scarcity makes them difficult to train against seasonal and day-night variations and thus less applicable in practice.

On the other hand, image retrieval is a much simpler task and performs well due to recent developments in image feature learning with CNNs ~\cite{krizhevsky2012imagenet, Simonyan2014, schroff2015facenet}. Such methods have provided promising feature encodings robust to various imaging and environment conditions while being well-suited for retrieval. Additionally, the large-scale dataset ImageNet~{\cite{deng2009imagenet}} has been instrumental in providing pre-trained models, which can be fine-tuned to other tasks, such as image-based localization, using transfer learning.
Finally, large datasets such as KITTI~{\cite{Geiger2012CVPR}} and the Oxford RobotCar~{\cite{maddern20171}} dataset have offered diverse training data for retrieval-based localization.

Despite the highlighted advantages, a drawback of state-of-the-art retrieval-based localization methods is that the learned features do not relate geometrically with each other. This may result in poor performance during localization if the reference images are sparsely located. Moreover, it also means that precise localization may not be possible with such a feature embedding. Although several methods have shown accurate localization by learning features from densely distributed images~\cite{Arandjelovic2014,Kim2017,Anoosheh2018}, this may not always be feasible due to high computational and memory requirements. We argue that learning features whose distances are directly proportional to the geometric counterpart in the map, results in more versatile and powerful features which also provide higher retrieval accuracy. Furthermore, such features are themselves of theoretical importance.

In this paper, we propose a method to do exactly so by adding a geometric cost to the visual cost of~\cite{schroff2015facenet}. Our new geometric loss ensures the proportionality relationship between geometric and feature distances while the visual cost uses the triplet loss~\cite{schroff2015facenet,weinberger2006distance} or its variants~\cite{chen2017beyond,angelina2018pointnetvlad} in order to tackle difficult false positives. Leveraging on the proposed learning mechanism, we consequently obtain features that are directly mappable. In other words, we demonstrate that for a given sub-sequence of query images, our learned features can be used to recover the corresponding trajectory without relying on any reference images.
Features of this kind may find their use in other applications apart from trajectory recovery such as map densification. We conduct experiments on several sequences of the Oxford RobotCar dataset~\cite{maddern20171} and the COLD indoor localization dataset~\cite{pronobis2009cold}. 
The experiments show that our features, whose distances are strongly correlated with the geometric distances, achieve significantly better retrieval accuracy. In practice, we obtain the proportionality relationship between geometric and feature distances when the images share visibility, resulting in an incomplete distance matrix. Despite that, we provide a method for recovering the original trajectory points solely from feature distances using the work of Dokmanic et al.~\cite{dokmanic2015euclidean}. We show promising results on the trajectory estimation, underlining the capabilities of geometrically mappable image features.

In summary, this paper introduces the following three contributions:
\begin{itemize}
 \item We introduce a novel loss function which is capable of learning features whose distances are locally proportional to their corresponding geometric distances.
 \item We show that our loss function significantly improves localization on two public datasets. 
 \item We demonstrate how mappable features can be used to approximately reconstruct an unknown map purely from feature distances.
\end{itemize}

\section{Related Work}
We briefly describe the related works for localization by image retrieval. As image retrieval requires highly descriptive and matchable features, most research in image retrieval has addressed the problem of learning good features relying on matching constraints. The triplet loss~\cite{schroff2015facenet,weinberger2006distance} has proved to be highly useful for image retrieval tasks. NetVLAD~\cite{arandjelovic2016netvlad} pools lower dimensional descriptive features for localization on top of VGGNet~\cite{Simonyan2014} using the triplet loss and has been influential in retrieval-based localization. \cite{Kim2017} train a network to further discard NetVLAD~\cite{arandjelovic2016netvlad} features that are irrelevant for localization. \cite{Gordo2017} improve on the training strategy with the triplet loss using the so-called Siamese architecture. Recent methods~\cite{angelina2018pointnetvlad,chen2017beyond} have proposed variants of the triplet loss to improve feature learning. A recent work~\cite{Duan_2019_CVPR} proposes a method to improve mining of triplets composing of hard negatives for training. A few works have addressed the problem of seasonal or day-night variations either by using 3D point clouds~\cite{angelina2018pointnetvlad} or by domain transfer~\cite{Anoosheh2018}. Others have proposed better or faster matching~\cite{Philbin2007,Stumm2015}, facilitating image retrieval. In the following sections we describe our method of improved feature learning based on the proposed geometric loss and show its implications for localization related tasks.

\begin{figure*}[t!]
\includegraphics[width=\textwidth]{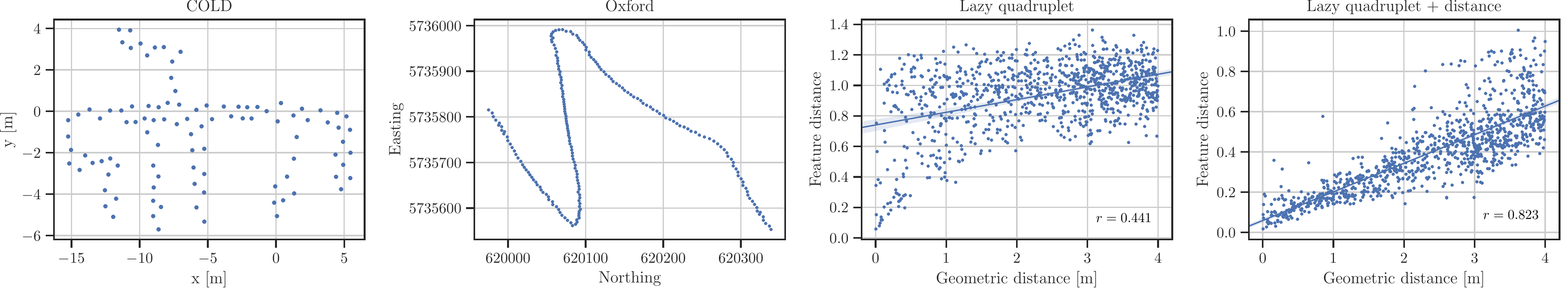}
\caption{Left two plots: Landmark locations obtained by greedy sampling with 100 landmarks for COLD Freiburg A and 200 landmarks for Oxford RobotCar. Right two images: 
Scatter plots of pairwise feature vs.~geometric distances between images from the COLD Freiburg (extended part A sunny sequence 3) after training with two different types of loss function---lazy quadruplet and lazy quadruplet + distance loss (one of ours). The plotted sequence was not used for training. It can be seen that the correlation between feature and geometric distances---which we exploit for mapping---is much better for our loss (Pearson`s correlation of 0.823) than for plain lazy quadruplet loss (Pearson`s correlation of 0.441).}
\label{fig:g_vs_f}
\end{figure*}

\section{Feature Learning}\label{sec:featureLearning}
Let us consider a set of given images $\{\mathcal{I}_i\}_{i=1}^n$ and their corresponding 2D locations $\{ \mathsf{x}_i\in \mathbb{R}^2 \}_{i=1}^n$. We assume that the location information is available together with images, either from cameras localized in the scene, or simply from sensor measurements such as GPS-tags\footnote{Although presented for 2D locations, our method can also be used when 3D locations are known.}. We wish to learn the image features $\{\mathsf{f}_i\}_{i=1}^n$ each representing the corresponding tuple in $\{(\mathcal{I}_i,\mathsf{x}_i)\}_{i=1}^n$. We further consider a set of image pairs $\mathcal{P} = \{\mathcal{P}_{ij}\} = \{(\mathcal{I}_i,\mathcal{I}_j)\}_{i,j=1}^n$, and define its positive subset $\mathcal{P}^+$ and negative subset $\mathcal{P}^-$ such that,
 {
 \medmuskip=0mu
\thinmuskip=0mu
\thickmuskip=0mu
\begin{equation}
 \mathcal{P}^+ = \{\mathcal{P}_{ij}: \,\, \mathbf{d}(\mathsf{x}_i,\mathsf{x}_j)\leq r_1 \}, \,\,\, \text{ and }\,\,\,  \mathcal{P}^- = \{\mathcal{P}_{ij}: \,\, \mathbf{d}(\mathsf{x}_i,\mathsf{x}_j)\geq r_2 \},\label{eq:pos_neg}
\end{equation}}
where $\mathbf{d}(.,.)$ is the distance function and $r_1,r_2$, with $r_1<r_2$, are the radii within and outside of which the image pairs are treated as positive and negative, respectively. For the positive pairs, we wish to learn image features which reflect the geometric distances. However, it may be unreasonable to expect the same for negative pairs. Therefore, given a set of images and their locations, the task of geometrically mappable feature learning aims at learning a mapping function $\mathcal{M}_\theta$, which obeys,
 {
 \medmuskip=0mu
\thinmuskip=0mu
\thickmuskip=0mu
 \begin{equation}\label{eq:conditionalFeature}
 \mathcal{M}_\theta:\mathcal{I}_i\rightarrow \mathsf{f}_i: 
 \begin{cases}
  \mathbf{d}(\mathsf{x}_i,\mathsf{x}_j) = \lambda \mathbf{d}(\mathsf{f}_i,\mathsf{f}_j), & \text{if}\ \mathcal{P}_{ij}\in\mathcal{P}^+, \\
  \mathbf{d}(\mathsf{f}_i,\mathsf{f}_j)> \mathbf{d}(\mathsf{f}_j,\mathsf{f}_l) +\alpha, \forall \mathcal{P}_{jl}\in\mathcal{P}^+ & \text{if\,} \mathcal{P}_{ij}\in\mathcal{P}^-,
 \end{cases}
 \end{equation}}
 where $\lambda >0$ is the proportionality constant, $\alpha$ is the so-called safety margin, and $\theta$ are the model parameters that we wish to learn from the data. In the following, we describe two different loss functions, derived from~\eqref{eq:conditionalFeature}, which will be minimized to learn the parameters $\theta$. 
\subsection{Visual-Geometric Loss}
We are interested on learning image features such that their distances reflect the real world geometric counterpart. This however may not be possible, especially for the images that are too far or do not even share any co-visibility. Therefore, in~\eqref{eq:conditionalFeature}, we impose the desired conditions only between images which are nearby in the geometric space. In fact, such consideration may not be sufficient due to many causes including, images with bad quality (e.g.~over or under exposed images and light glares) or noisy and incorrect image locations and angles. Therefore, we make use of the robust regression technique and design the visual-geometric proportionality loss function as follows:
\begin{equation}\label{eq:geometricLoss}
 \mathcal{L}_{VG} = \sum_{\mathcal{P}_{ij}\in\mathcal{P}^+}{\rho(\mathbf{d}(\mathsf{x}_i,\mathsf{x}_j) - \lambda \mathbf{d}(\mathsf{f}_i,\mathsf{f}_j))},
\end{equation}
where, $\rho(.)$ is the robust Huber loss function. 
\subsection{Negative Visual Loss}
When the feature distances cannot be directly related to the geometric distances, we wish to keep the feature distances of the negative pair (i.e. two images far apart) at least $\alpha$ farther than that of the positive pair, as stated by the second condition in~\eqref{eq:conditionalFeature}. To ensure this property, we make use of the hinge loss function given as follows for positive and negative pairs,
\begin{equation}
 \mathcal{H}(\mathcal{P}_{ij},\mathcal{P}_{jl}) = \text{max}\{0, \mathbf{d}(\mathsf{f}_j,\mathsf{f}_l)- \mathbf{d}(\mathsf{f}_i,\mathsf{f}_j) +\alpha\}.
\end{equation}
The loss function imposing margin $\alpha$ on the distances between positive and negative image pairs is given by,
\begin{equation}\label{eq:triptletLoss}
 \mathcal{L}_{NV} = \sum_{_{\mathcal{P}_{ij}\in \mathcal{P}^+, \mathcal{P}_{jl}\in \mathcal{P}^-}}{\mathcal{H}(\mathcal{P}_{ij},\mathcal{P}_{jl})}.
\end{equation}
In practice, one can use several variations of~\eqref{eq:triptletLoss}. Three of these variations include lazy triplet~\cite{angelina2018pointnetvlad}, quadruplet~\cite{chen2017beyond}, and lazy quadruplet~\cite{angelina2018pointnetvlad} loss functions. We seek for the model parameters $\theta$ by jointly minimizing the loss functions~\eqref{eq:geometricLoss} and \eqref{eq:triptletLoss}. Please refer to Section~\ref{sec:implementationDetails} for the details about several variations, and the choice of hyperparameter between losses $\mathcal{L}_{VG} $ and $\mathcal{L}_{NV}$. 

\section{Localization and Mapping}
In this section, we show how the feature distances, obtained using the method presented in Section~\ref{sec:featureLearning}, can be used to better localize and map the camera trajectories, demonstrating that our features are indeed mappable.
Let $\mathsf{D}\in\mathbb{R}^{m\times n}$ be a Euclidean Distance Matrix (EDM)~\cite{dokmanic2015euclidean}, whose entries $d_{ij}$ are the squared feature distances between images $\mathcal{I}_i$ and $\mathcal{I}_j$.
Without loss of generality, we consider that both $\{\mathcal{I}_i\}$ and $\{\mathcal{I}_j\}$ are query images\footnote{The problem becomes simpler if either $\{\mathcal{I}_i\}$ or $\{\mathcal{I}_j\}$ are reference images with known locations. Under such circumstances, one can easily estimate the locations of query images. 
Details of this case will not be discussed in this paper. Please refer to~\cite{dokmanic2015euclidean} for more information.}. Therefore, we are now interested in the case when $\mathsf{D}\in\mathbb{R}^{n\times n}$ from which we wish to recover the 2D locations of images, under the assumptions of Section~\ref{sec:featureLearning} and in the presence of noise. 
We make use of so-called multidimensional scaling (MDS)---the problem of finding the best point set representation for a given set of distances~\cite{dokmanic2015euclidean}.
MDS, however, requires an EDM with little noise and no missing data. In the following, we describe the method we follow to recover EDM and image locations using MDS.
\subsection{EDM Recovery}
Recall from~\eqref{eq:conditionalFeature} that we cannot rely on distances $d_{ij}\geq r_1$. Let $\mathsf{M}\in\mathbb{R}^{n\times n}$ be a mask, whose entries $m_{ij}=1$ if $d_{ij}\leq r_1$ and $m_{ij}=0$ otherwise. Now, we wish to recover the full EDM from $\mathsf{M}\odot \mathsf{D}$. Let $\mathsf{X} = [\mathsf{x}_1\,, \mathsf{x}_2\,,\ldots, \mathsf{x}_n]\in \mathbb{R}^{2 \times n}$ be the desired 2D coordinates of the image locations and the Gram matrix $\mathsf{G}=\mathsf{X^TX}$. By definition, matrix $\mathsf{G}$ is of at most rank 2. Furthermore, we also define an operator $\mathcal{K}(\mathsf{G})$, similar to the EDM, as
\begin{equation}
 \mathcal{K}(\mathsf{G}) := \text{diag}(\mathsf{G})\mathbf{1}^\mathsf{T} -2\mathsf{G} +\mathbf{1}\text{diag}(\mathsf{G})^\mathsf{T}. 
\end{equation}
Now, the task of EDM recovery, in the presence of noise and missing data, can be formulated as finding the Gram matrix $\mathsf{G}$ by solving the following rank-constrained semi-definite program (SDP)~\cite{dokmanic2015euclidean}, 
\begin{align}
\label{eq:optimizationSDP}
\begin{split}
\underset{\mathsf{G}}{\text{minimize}}\quad &\norm{\mathsf{M}\odot(\mathsf{D}-\mathcal{K}(\mathsf{G}))}_{F}^2,\\
\text{subject to}\quad &\text{rank}(\mathsf{G})\leq 2, \mathsf{G}\succeq 0, \mathsf{G}\mathbf{1}=0.
\end{split}
\end{align}
The rank constraint of problem~\eqref{eq:optimizationSDP} is non-convex. However, it is a common practice in EDM to drop this rank~\cite{boyd1994linear}. In fact, a tighter relaxation of this problem can be formulated in the Lagrangian form, avoided here for brevity, under the same spirit. Please, refer to~\cite{dokmanic2015euclidean} for tighter SDP relaxation. 
\subsection{Image Location Recovery}
Once the Gram matrix $\mathsf{G}$ is recovered, the image locations $\{\mathsf{x}_i\}$ can be found by using the Eigen Value Decomposition (EVD) of $\mathsf{G}$. Let the EVD be $\mathsf{G}=\mathsf{U}\mathsf{\Lambda}\mathsf{U}^T$. If $\Tilde{\mathsf{\Lambda}}$ is the truncated $\mathsf{\Lambda}$, except for the largest two eigenvalues, the recovered image locations are given by $\Tilde{\mathsf{X}} = \Tilde{\mathsf{\Lambda}}^{1/2}\mathsf{U^T}$.
In fact, it is straightforward to see from the definition of $\mathsf{G}:=\mathsf{X^TX}$ that such recovery is optimal. This process of recovery is called the classical MDS, which makes a particular coordinate choice of $\mathsf{x}_1$ being the origin. Classical MDS, can be improved by iterative refinement methods such as SMACOF~\cite{Leeuw1977} or its variants. In the case of known reference points, one may obtain a better recovery simply by constraining the entries of matrix $\mathsf{G}$. Alternatively, other techniques for location (also called ``point set" in the literature) recovery do exist. Exploration of these techniques is left for future research.

\begin{figure*}[t]
\includegraphics[width=\textwidth]{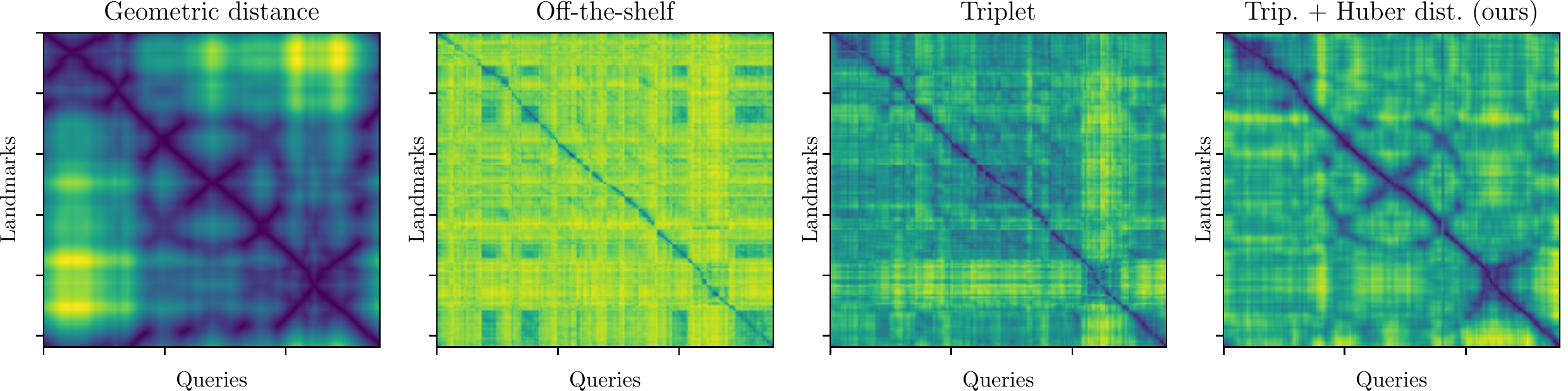}
\caption{Feature distance matrices between a cloudy reference and a cloudy query sequence from COLD Freiburg A. The underlying features are obtained with a network that was not retrained for localization (off-the-shelf), triplet loss, and our combination of triplet and Huber loss, respectively. As a reference, the ground truth geometric distances are also provided. Ideally, a feature distance matrix should be proportional to the geometric distance matrix.}
\label{fig:d_matrix}
\end{figure*}

\section{Implementation Details}
\label{sec:implementationDetails}
\subsection{Datasets and Baselines}
We conduct experiments on two publicly available real datasets---Oxford RobotCar~\cite{maddern20171} and the COLD database~\cite{pronobis2009cold}. 
From the Oxford RobotCar dataset we use the first 1.25km of the sequences 2014.11.18 13:20:12 (sun, clouds), 2014.12.17 18:18:43 (night, rain), 2015.02.03 08:45:10 (snow), and 2014.12.02 15:30:08 (overcast) for training.
We set aside 2015.02.13 09:16:26 (overcast), 2014.12.16 18:44:24 (night), 2015.10.29 12:18:17 (rain) for testing. This results in 15006 images for training and 12329 testing images.

Form the COLD database we mostly use Freiburg region A, where we consider all sequences in part A, setting aside the third runs (normal and extended) for testing, using the remaining sequences for training our networks.
This adds up to 29237 training and 12831 testing images. 

To evaluate the generalization capability of features trained on the COLD Freiburg A training data, we use three different regions of the COLD database, Freiburg A (extended), Freiburg B and Saarbrucken B. For these experiments, when evaluating on COLD region A, we use the second extended sunny sequence (the first one is missing a room) as a reference and the third extended sunny, cloudy, and night sequences as queries. For COLD Freiburg B, we use the first cloudy sequence as reference and the second cloudy and sunny sequences as queries. There are no night sequences for COLD Freiburg part B. For COLD Saarbrucken B, the reference is the first cloudy sequence and the second cloudy, night, and sunny sequences are used as queries.

We compare our method against five existing methods, namely off-the-shelf ImageNet features without localization fine-tuning~{\cite{Simonyan2014}}, features fine-tuned with triplet~{\cite{arandjelovic2016netvlad}},
quadruplet~{\cite{chen2017beyond}},
lazy triplet~{\cite{angelina2018pointnetvlad}}, and
lazy quadruplet~{\cite{angelina2018pointnetvlad}} loss.

\subsection{Network Architecture and Weights}
We use a VGG-16~\cite{Simonyan2014} network cropped at the last convolutional layer and extend it with a NetVLAD~\cite{arandjelovic2016netvlad} layer as implemented by~\cite{cieslewski2018data}, initializing the network with off-the-shelf ImageNet~\cite{deng2009imagenet} classification weights, i.e.\ weights that have not yet been retrained for localization. The weights for the NetVLAD layer are calculated using 30'000 images from Pittsburgh 250k \cite{torii2013visual}.

\begin{figure*}[t!]
\includegraphics[width=\textwidth]{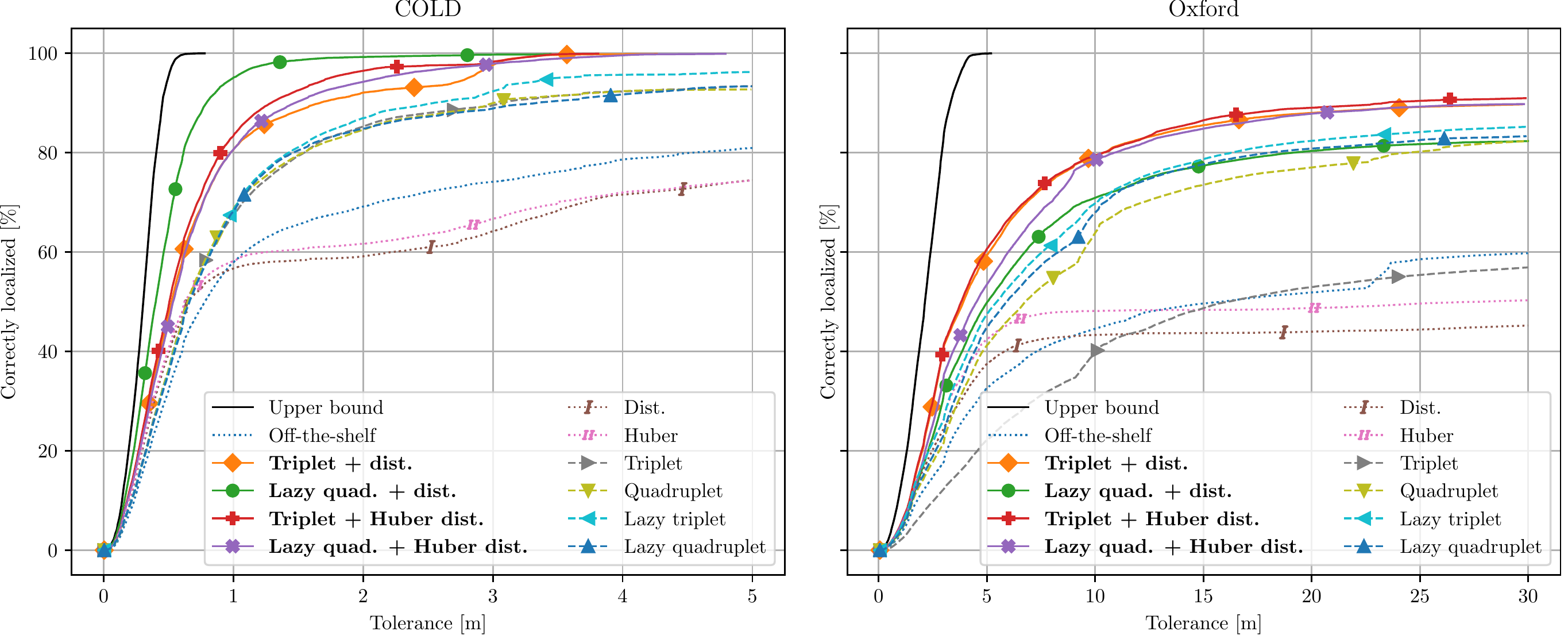}
\caption{Percentage of correctly localized images for varying distance tolerance. The upper bound is given by the distance of each query to its geometrically closest reference image and is plotted in solid black. Off-the-shelf features (dotted blue) directly use weights trained for ImageNet~\cite{deng2009imagenet} classification without retraining for localization. Pure visual-geometric losses are drawn in dotted purple and pink while triplet-like losses are shown with dashed lines. Our methods (solid lines with markers), combine these two loss types and achieve the best results for both datasets.} 
\label{fig:roc}
\end{figure*}

\subsection{Training}
We use the training script from~\cite{angelina2018pointnetvlad}. 
Note that this training script is designed to train PointNet~{\cite{qi2017pointnet}} with a NetVLAD{~\cite{arandjelovic2016netvlad}} layer for point cloud based localization.
To use the training script for images instead, we replace all point cloud specific loaders and containers with their image counterparts. We replace the network to be trained with the VGG-16 and NetVLAD architecture implementation of~{\cite{cieslewski2018data}}.
Furthermore, we use the training parameters specified in~\cite{arandjelovic2016netvlad}, reducing the learning rate to 0.000001, the number of queries per batch to two, and the number of positives and negatives per query to six each.
With this smaller query size it becomes possible to train VGG-16 in its entirety and not only down to conv5 layer as it is done by~\cite{arandjelovic2016netvlad}.

During training, for each epoch, we iterate over all training images.
For each image, we sample positives from within a radius of $r_1$, and negatives that are at least $r_2$ distance away.
We set $r_1$ to 1 meter for COLD Freiburg and, in accordance with~\cite{arandjelovic2016netvlad}, to 10 meters for Oxford RobotCar. 
$r_2$ is set to 4 meters and 25 meters respectively. 
Additionally, for COLD Freiburg, we exclude images with a difference in yaw angle larger than 30 degrees. This minimizes the problem of lack of co-visibility between images. The angle restriction was usually respected in Oxford RobotCar dataset, therefore the angle based filtering was not performed.
For each image, half of the negatives are chosen via hard negative mining. 
We update our feature cache for hard negative mining every 400 iterations for COLD Freiburg and every 1000 iterations for Oxford RobotCar.

We train the COLD Freiburg networks for roughly 48 hours, which equates to one epoch. 
The Oxford RobotCar networks are trained for 24 hours, resulting in two epochs.
The reason for such fast convergence is that after one epoch, the network will have seen every image an average of 13 times, once as an anchor and 12 times as a positive or negative.
Hard negative mining also contributes to faster convergence by selecting relevant tuples for training.
We only train one model per loss function and dataset.

\subsection{Losses}
We run experiments with two types of visual-geometric loss, Huber \eqref{eq:geometricLoss} and a simple distance loss, which replaces $\rho$ in \eqref{eq:geometricLoss} with a squared Euclidean distance.
The $\mathbf{d}(\mathsf{x}_i,\mathsf{x}_j)$ and $\mathbf{d}(\mathsf{f}_i,\mathsf{f}_j)$ in \eqref{eq:geometricLoss} are also squared Euclidean distances and $\lambda$ is chosen based on $r_1$ and the maximum off-the-shelf pairwise feature distance between training images.
\cite{arandjelovic2016netvlad} adapted the commonly used triplet loss \cite{weinberger2006distance,wang2014learning,schroff2015facenet,schultz2004learning} for weakly supervised training. We use this adapted triplet loss, its quadruplet extension, and the lazy triplet and lazy quadruplet versions introduced by \cite{angelina2018pointnetvlad}. For all four losses, we use the implementation from \cite{angelina2018pointnetvlad}. 
Our final loss functions are given by $\mathcal{L}= \mathcal{L}_{NV} + \gamma\mathcal{L}_{VG}$, where $\gamma$ is a scaling factor. Auxiliary experiments show, that the training process is robust with respect to parameter $\gamma$, which has been set empirically to $0.5$ for all experiments in this paper.

\begin{table}[tb]
\begin{center}
\caption{Percentage of correctly localized images (bold: our methods, italic: best method per column).}
\resizebox{\columnwidth}{!}{%
\begin{tabular}{lcccc}
 & \multicolumn{2}{c}{COLD} & \multicolumn{2}{c}{Oxford} \\ \hline
Loss & 1m & 4m & 10m & 25m \\ \hline
Off-the-shelf~\cite{Simonyan2014} & 57.92 & 78.65 & 44.56 & 58.55 \\
Dist. & 56.71 & 71.56 & 43.31 & 44.39 \\
Huber dist. & 58.23 & 72.00 & 48.15 & 49.65 \\
Triplet~\cite{arandjelovic2016netvlad} & 67.45 & 92.32 & 39.67 & 55.40 \\
Quadruplet~\cite{chen2017beyond} & 68.76 & 92.40 & 63.92 & 80.21 \\
Lazy triplet~\cite{angelina2018pointnetvlad} & 68.55 & 95.69 & 69.85 & 84.19 \\
Lazy quadruplet~\cite{angelina2018pointnetvlad} & 68.30 & 91.78 & 67.94 & 82.52\\
\textbf{Triplet + dist.} & 80.66 & 99.81 & 79.33 & 89.23 \\
\textbf{Lazy quad. + dist.} & \textit{95.10 }& 99.85 & 70.91 & 81.73 \\
\textbf{Triplet + huber dist}. & 83.47 & \textit{99.94} &\textit{ 79.46} & \textit{90.44 }\\
\textbf{Lazy quad. + huber dist.} & 80.63 & 99.61 & 78.51 & 89.28 \\
\hline%
\end{tabular}
}
\label{tab:localization_accuracy}
\end{center}
\end{table}

\subsection{Reference Landmark Selection}
The larger the number of reference landmark images in a database for image-based localization, the larger the memory requirements and the slower the retrieval of the closest match to an incoming query image.
For this reason, the number of landmarks is usually kept small. 
In this paper we use two simple methods for selecting reference landmarks from one or more sequences of potential reference images.

For most experiments we use greedy sampling, where the first landmark is selected randomly and every subsequent landmark is selected as the image which has the largest geometric distance to all images in the set of already selected landmarks.
This is repeated until the desired number of landmarks is reached.
Fig.~{\ref{fig:g_vs_f}} shows the landmark locations obtained with greedy sampling that were used for the quantitative evaluations reported in Fig.~{\ref{fig:roc}} and Tab.~{\ref{tab:localization_accuracy}}.

When comparing the localization performance on different regions with different sizes, it is no longer adequate to fix the number of reference landmarks.
Instead, we choose one reference sequence.
The first landmark is the first image in the sequence.
Thereafter, we iterate through the reference sequence and add each image that is at least $r_{\mathrm{LM}}$ away from the last selected landmark to the list of landmarks. 

\begin{figure*}[t!]
\includegraphics[width=\textwidth, trim={0 1.8mm 0 0}, clip]{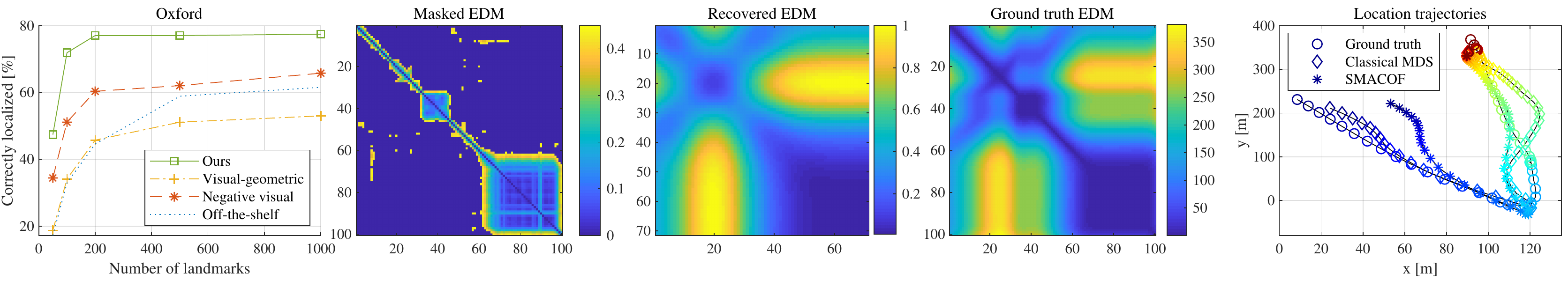}
\caption{First plot from the left: Localization accuracy for different numbers of landmarks on Oxford. 
 Second through fourth plots: Masked, recovered (using SDP), and ground truth distance matrices. Note that the different scales are due to the proportionality factor $\lambda$ in \eqref{eq:geometricLoss}. Right plot: Ground truth and reconstructed image trajectories obtained without using reference image locations. Same marker colors indicate corresponding points in the three trajectories.}
\label{fig:mapRecovery}
\end{figure*}

\begin{figure*}[p]
 \centering
	\includegraphics[width=0.82\textwidth]{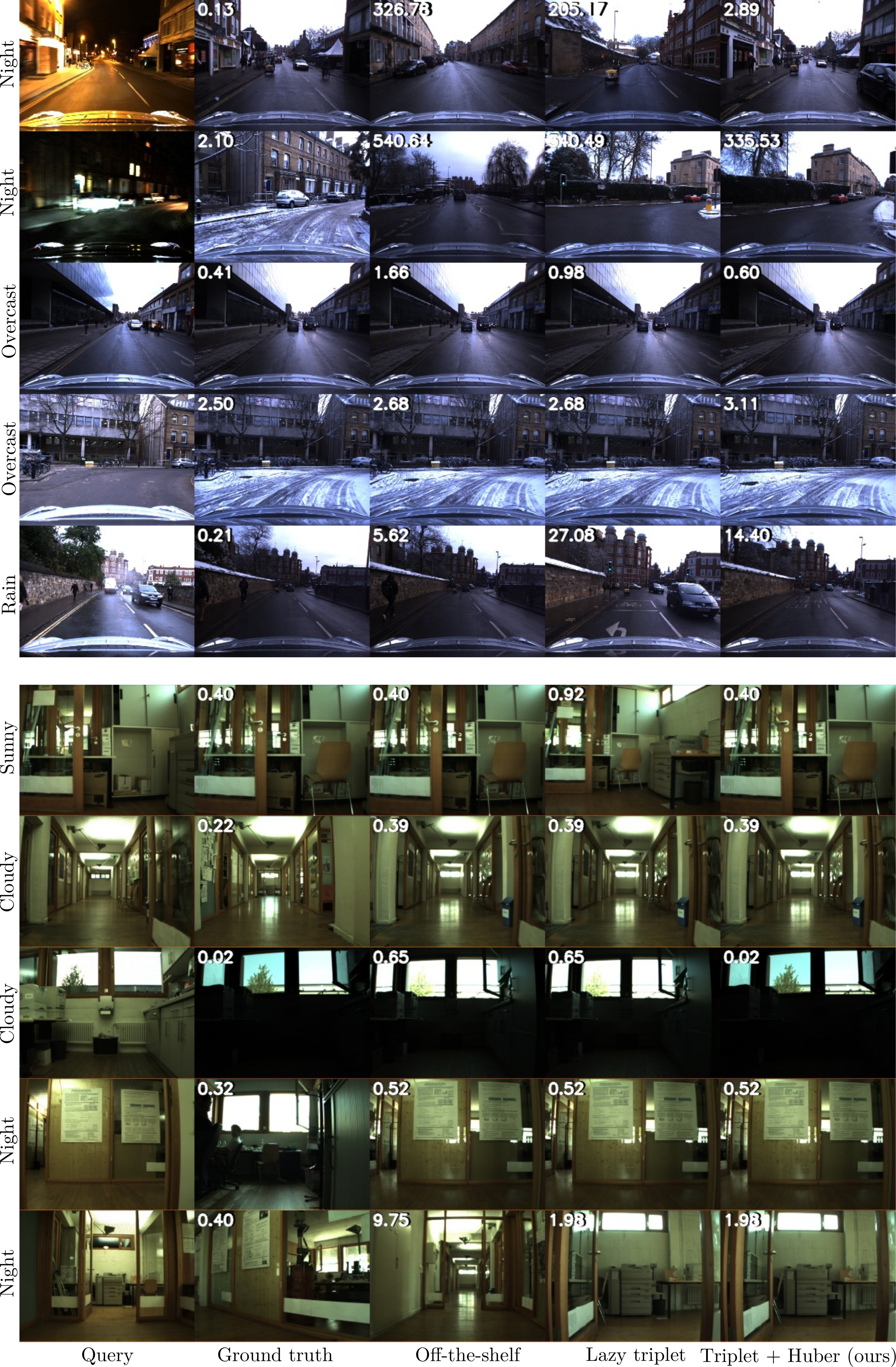}
 \caption{Visual results for localizing random query images taken under various conditions against a set of 1000 greedily sampled snowy reference images from the Oxford RobotCar dataset (top) and against 100 greedily sampled sunny reference images from the COLD Freiburg dataset (bottom). The ground truth is the geometrically closest landmark image. The numbers indicate the distance between true query locations and the retrieved reference locations in meters.}
 \label{fig:visual_results}
\end{figure*}

\section{Experiments}
\subsection{Visual-Geometric Proportionality} 
We aim to train features for which the distances of two nearby images become proportional to their geometric distances.
We verify this property by plotting pairwise feature distances against the corresponding geometric distances. 
Fig.~\ref{fig:g_vs_f} shows such plots for lazy quadruplet, and our combination of distance and lazy quadruplet loss. 
The plots were generated using the features and locations from the extended sunny testing sequence of the COLD Freiburg dataset.
Similar results can be observed for cloudy, night, and mixed sequences.
Looking at the Pearson`s correlation reported in Fig.~{\ref{fig:g_vs_f}}, it can be seen that adding the visual-geometric loss significantly increases the correlation between feature and geometric distances. 

The first image in Fig.~{\ref{fig:d_matrix}} shows the geometric distance matrix between a cloudy reference and a cloudy query sequence from COLD Freiburg A.
The other images show the corresponding feature distance matrices, obtained with different types of features---off-the-shelf, triplet, and our combination of triplet and Huber.
Note that the diagonal blue demarcation in the geometric distance matrix is also visible in the feature distance matrices of our features, showing that the geometrically close images are also close in feature space.
Please refer to our supplementary video for an animated illustration of the visual-geometric proportionality. 

\subsection{Quantitative Localization Performance}
\label{sec:quantitative_analysis}
Fig.~\ref{fig:roc} shows the percentage of correctly localized images for varying distance tolerance.
For the COLD Freiburg dataset, these values are calculated by matching all images from all testing sequences (sunny, cloudy, and night) to their closest reference landmarks out of 100 geometrically greedily sampled training images taken from all conditions.
For the Oxford RobotCar dataset, we use all images form the night, rain and cloudy testing sequences and match them against 200 landmarks sampled from all training conditions (sun/clouds, snow, night/rain, and overcast). Additionally, Tab.~\ref{tab:localization_accuracy} states the exact localization accuracy at $r_1$ and $r_2$, i.e.\ the maximum positive and minimum negative distances during training (1m/4m for COLD Freiburg and 10m/25m for Oxford RobotCar). The proposed methods, i.e.\ combinations of a visual-geometric loss (dist.\ or Huber dist.) combined with a variation of triplet loss, are emphasized in bold font and the best accuracy for each location and threshold is given in italic.
Fig.~\ref{fig:roc} and Tab.~\ref{tab:localization_accuracy} clearly show the better localization performance of the proposed combined loss functions in comparison to pure visual-geometric and triplet-like loss functions.
Auxiliary experiments show that using Huber instead of a squared Euclidean distance loss results in less deviation in performance between different instances of the same training process.
The first plot in Fig.~{\ref{fig:mapRecovery}} also shows that superiority of our loss persists across different numbers of reference landmarks.

\subsection{Generalization Performance}
We test the ability of our descriptors to generalize to new places and conditions by evaluating the models trained on COLD Freiburg region A on the geographically disjoint COLD Freiburg region B and on COLD Saarbrucken part B, which was acquired with a different robot.

For a reference landmark distance of $r_{\mathrm{LM}}=1\mathrm{m}$ the percentage of images for which the top-1 retrieved landmark lies within 1m of the query location is, on average, 87.02\% on Freiburg A, 77.65\% on Freiburg B, and 69.92\% on Saarbrucken B when using features that were only trained with a negative visual loss (i.e.~triplet, quadruplet, etc.). Adding a visual-geometric loss, significantly increases the performance on Freiburg A to 94.79\%, and achieves results similar to pure negative visual loss on Freiburg B (76.18\%) and Saarbrucken B (67.99\%).
Increasing $r_{\mathrm{LM}}$ to 1.25m, i.e.~a value slightly larger than the original maximal positive radius, will change these numbers to 82.42\%, 73.52\%, and 60.88\% for pure negative visual losses and 94.34\%, 76.47\%, and 60.92\% for the combinations of negative visual and visual-geometric losses.

These results show that the performance of all descriptors suffer in the presence of domain shifts. 
It is therefore always best to train directly for the region where one wishes to localize new images.
If such local data is available, using mappable features as proposed in our paper, will yield significantly better results locally while generalizing just as well as triplet loss or other negative visual losses to new conditions.
This allows us to gracefully handle extensions and adjustments to the reference region.
E.g.~remodeling or adding some rooms does not require retraining.
Moreover, the mappability characteristics of our descriptors allows to find an approximate location for new landmark candidates---potentially sourced from the pool of submitted queries---with respect to the known map without the need for 3D reconstruction.

\subsection{Qualitative Localization Performance}
In Fig.~\ref{fig:visual_results} we localize query images (first column from the left) from all testing conditions against the snowy Oxford training sequence and the first extended sunny training sequence from the COLD dataset, using different types of features.
The right most column shows the reference images retrieved with our combination of triplet and Huber loss. 
It can be seen that our method is able to learn features which correctly localize query images taken under very different conditions from the reference set, such as the night image in the top row. 
Please refer to our supplementary video for an animated illustration of the localization performance of an entire image sequence.

\begin{table}[t!]
\centering
\caption{Localization time and standard deviation.}
\resizebox{\columnwidth}{!}{%
\begin{tabular}{lcc}
 & COLD & Oxford \\ \hline
GPU query feature inference & $7.13\pm0.09\text{ms}$ & $6.20\pm0.17\text{ms}$ \\ 
CPU top-1 landmark retrieval & $4.65\pm1.87\text{ms}$ & $6.81\pm0.30\text{ms}$ \\ \hline
Total & $11.78\pm1.96\text{ms}$ & $13.01\pm0.13\text{ms}$ \\ \hline
\end{tabular}%
}
\label{tab:time}
\end{table}

\subsection{Localization Speed}
Tab.~\ref{tab:time} shows the average time it takes to localize one query image in the setting described in Section~\ref{sec:quantitative_analysis}. The times are obtained using a GeForce GTX 1080 Ti GPU and an Intel Core i7-9700K CPU @ 3.60GHz with 32GB RAM.
Note that the localization improvements shown in Fig.~\ref{fig:roc} and Tab.~\ref{tab:localization_accuracy} are based on simple top-1 retrieval and do not leverage map recovery. Localization speed is, therefore, the same for all loss functions. Retrieval is done by querying a precalculated KD-tree containing all landmark features. We use 100 landmarks for COLD and 200 for Oxford. This is why retrieval is faster for COLD. The faster feature inference speed for Oxford is due to the smaller image size (240$\times$180) used for our experiments.

\subsection{Map Recovery}
Using the masked feature distance matrix, we first recover the missing entries for a sequence of Oxford dataset, which is shown in the Fig.~\ref{fig:mapRecovery}. In the same figure, we also show the ground truth distance matrix as a reference. We further perform image location recovery for the same sequence of images, without using any information about the reference images. The feature distances are computed only among query images, followed by EDM completion (using SDP~\cite{dokmanic2015euclidean}) and locations recovery using MDS and SMACOF. The recovered trajectories after alignment are shown in Fig.~\ref{fig:mapRecovery} (right). The ground truth trajectory has a length of 672m and the average RMSE over ten repetitions is 31.37m (4.67\%) for classical MDS and 38.77m (5.77\%) for SMACOF.

\section{Conclusion}
\label{sec:conclusion}
We introduce a class of loss functions which are able to learn visually representative and geometrically meaningful image features. More precisely, the proposed feature distances are locally proportional to their corresponding geometric distances. We show that our learned features significantly outperform the state of the art in terms of image-retrieval-based localization performance. 
This is because by transferring the distance metric from geometric space to the feature embedding during learning, we are able to obtain features that change smoothly with location. More  theoretical analysis in regard to better generalization still remains to be explored as a future work.
To further highlight the capabilities of our features, we demonstrate that the feature distances can be used to reconstruct the geometry of query sequences, even without the reference images.
The proposed method is generic and can be extended for the case of 3D landmarks in a straightforward manner.
The features learned in this paper can be used for fast and accurate condition invariant localization, and also potentially for map densification and updates. 


\bibliographystyle{IEEEtran}
\bibliography{IEEEabrv,clean_bib}
\clearpage
\end{document}